\documentclass[letterpaper, 10 pt, conference]{ieeeconf}  

\IEEEoverridecommandlockouts                              

\usepackage{graphics} 
\usepackage{epsfig} 
\usepackage{times} 
\usepackage{amsmath} 
\usepackage{amssymb}  
\usepackage{subcaption}
\usepackage{hyperref}
\usepackage[table]{xcolor}

\usepackage{pgfplots}

\pgfplotsset{compat=1.17}
\usetikzlibrary{pgfplots.groupplots}

\newcommand{\famp}{FA-ProDMP}
\newcommand{\cvec}[1]{\boldsymbol{{#1}}}
\newcommand{\cmat}[1]{\boldsymbol{\mathrm{#1}}}

\title{Use the Force, Bot! - Force-Aware ProDMP with Event-Based Replanning
}

\author{Paul Werner L\"odige$^{* 1}$, Maximilian Xiling Li$^{* 1}$ and Rudolf Lioutikov$^{1}$
\thanks{$^{1}$Intuitive Robots Lab, Karlsruhe Institute of Technology, Germany
        {\tt\small \{maximilian.li, lioutikov\}@kit.edu}}%
\thanks{$^{*}$Equal contribution}%
\thanks{This work was funded by the Deutsche Forschungsgemeinschaft (DFG, German
Research Foundation) – 448648559}
}

\begin{document}

\maketitle
\thispagestyle{empty}
\pagestyle{empty}

\begin{abstract}
Movement Primitives (MPs) are a well-established method for representing and generating modular robot trajectories. This work presents \famp{}, a novel approach that introduces force awareness to Probabilistic Dynamic Movement Primitives (ProDMP). \famp{} adapts trajectories during runtime to account for measured and desired forces, offering smooth trajectories and capturing position and force correlations across multiple demonstrations. \famp{}s support multiple axes of force, making them agnostic to Cartesian or joint space control.
This versatility makes \famp{} a valuable tool for learning contact rich manipulation tasks, such power plug insertion.
To reliably evaluate \famp{}, this work additionally introduces a modular, 3D printed task suite called POEMPEL, inspired by the popular \textit{Lego Technic} pins. POEMPEL mimics industrial peg-in-hole assembly tasks with force requirements and offers multiple parameters of adjustment, such as position, orientation and plug stiffness level, thereby varying the direction and amount of required forces.
Our experiments demonstrate that \famp{} outperforms other MP formulations on the POEMPEL setup and a electrical power plug insertion task, thanks to its replanning capabilities based on measured forces. These findings highlight how \famp{} enhances the performance of robotic systems in contact-rich manipulation tasks.
\end{abstract}
\section{Introduction}
\label{sec:intro}
Movement Primitives (MPs) are a prominent tool for representing and generating robot trajectories. They have gained much popularity in imitation and reinforcement learning \cite{otto23blackbox, Franzese2023InteractiveImitation}, due to their compact parameterization and flexibility. MPs have solved many tasks, including surgical robotic tasks \cite{Scheikl2024Movementprimitivediffusion}, Tabletennis \cite{Li2023Curriculumbasedimitation, blessing2024information} or Ball-Throwing \cite{Ude2010TaskSpecificGeneralization, Cohen2021Motion}.

MPs enable the creation of complex robot skills by combining and concatenating basic movement elements. They can be adapted to new target positions and velocities or via points \cite{Zhou2019Learningviapoint}.
Probabilistic approaches, such as Probabilistic Movement Primitives (ProMP) \cite{Paraschos2013Probabilisticmovementprimitives}, estimate the statistics of a distribution of trajectories, e.g. a set of expert demonstrations. These methods capture  correlations over time and across multiple Degrees of Freedom (DoF). ProMP trajectories can be combined to compose complex behaviors by using probabilistic operations \cite{Paraschos2017Priorization}.

Methods based on dynamical systems, such as Dynamic Movement Primitives (DMP) \cite{Schaal2006Dynamicmovementprimitives}, guarantee that a generated trajectory starts with the current robot position and velocity. This guarantee enables smooth trajectory replanning, i.e. changing the target trajectory by adapting the MP parameters during movement execution. Replanning allows DMPs to react to changes in the execution environment, e.g., in a human-robot interaction setting \cite{amor2014interaction, Zhou2016LearningForceAdaptation}.

Probabilistic Dynamic Movement Primitives (ProDMP) \cite{Li2023ProDMPUnifiedPerspective} unify both approaches and combines their strengths. ProDMP allow for smooth replanning during movement execution and capture the correlations of a set of demonstrations. But until now, ProDMP did not consider the force profile during task execution, limiting their potential applications.
Especially contact rich manipulation tasks, such as pen writing \cite{Steinmetz2015Simultaneouskinestheticteaching}, cutting vegetables \cite{Lioutikov2016LearningManipulationSequencing}, unscrewing a light-bulb \cite{Manschitz2016Probabilisticdecompositionsequential} or surface wiping \cite{Do2014LearnWipe, Wang2023ALProMPForce} require the movement follow a specific force profile.

\begin{figure}
    \centering
    \includegraphics[width=0.5\linewidth]{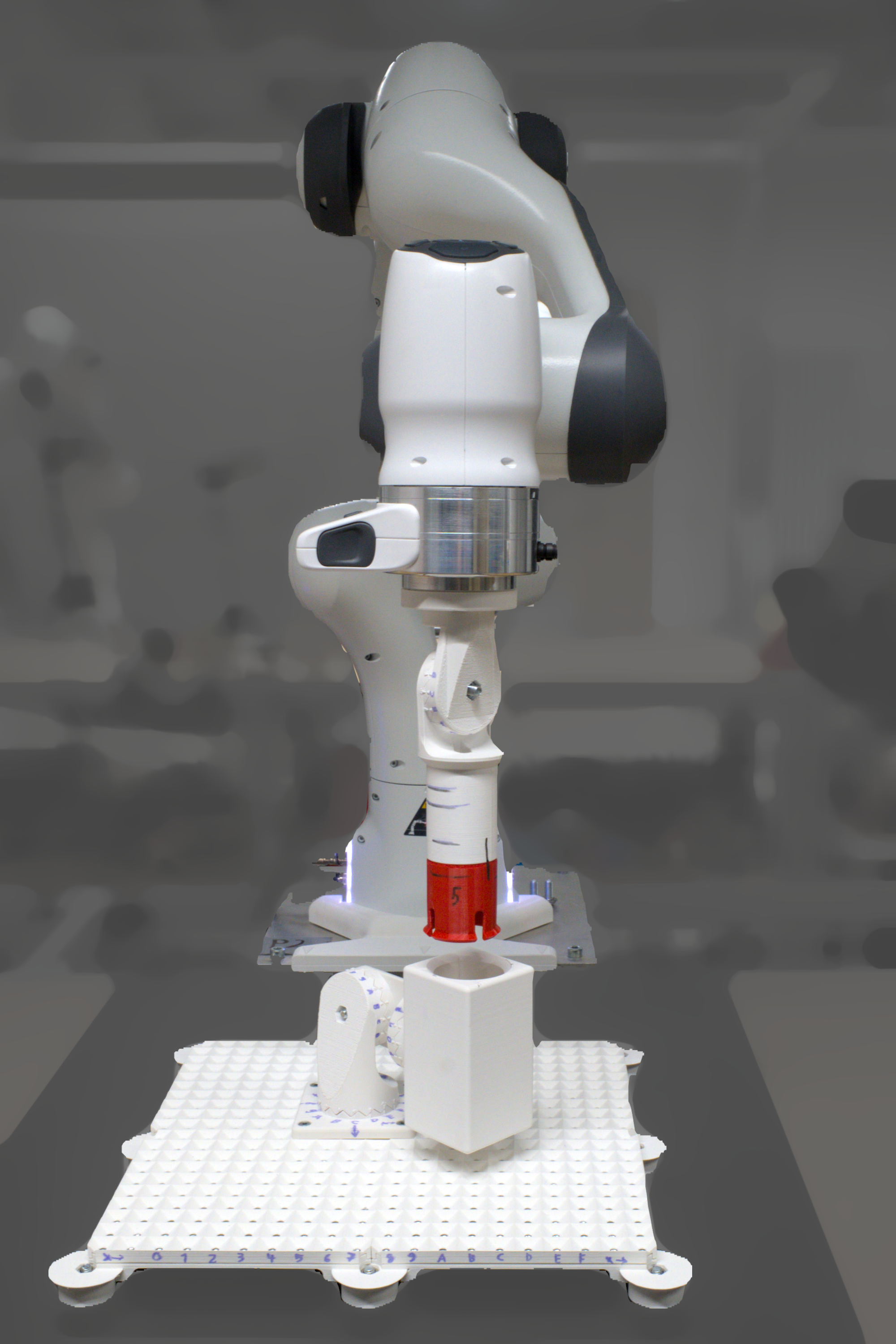}
    \caption{POEMPEL task suite mounted on a Franka Emika 7-DoF robot arm. All parts are indexed and identifiable to ensure repeatability across attempts and labs. The red tip of the end-effector mounted plug is made from a flexible TPU material. The tip can be exchanged to others with different construction parameters, resulting in different stiffness levels and force requirements.}
    \label{fig:poempel}
\end{figure}

The main contributions of this paper are as follows:
\begin{itemize}
    \item A new MP representation called Force-Aware Probabilistic Dynamic Movement Primitives (\famp), enabling ProDMP to capture force profiles from demonstrations and to be conditioned to enact desired forces.
    \item Event-based replanning for \famp, which adapts the MP trajectory during execution if measured forces exceed an expected threshold. This mechanism leverages the compositionality of ProDMP to adapt both position trajectory and desired forces.
    \item A new task suite named POEMPEL, mimicking industrial assembly tasks (ref. \autoref{fig:poempel}). POEMPEL is adjustable with multiple, indexed DoF in order to create tasks with varying force requirements in both strength and direction. It can be easily reproduced via 3D printing\footnote{Please find POEMPEL instructions at \href{https://intuitive-robots.github.io/poempel/}{https://intuitive-robots.github.io/poempel/}}.
    \item A comparison of \famp{}'s task performance on the POEMPEL setup and an electrical plug insertion task with other MP formulations such as DMP, ProMP, ProDMP and direct replay using a Cartesian impedance controller.
\end{itemize}
\section{Related Work}
\label{sec:related}
\textbf{Force Aware Movement Primitives.}
Prior work aimed to integrate force profiles into movement primitives. For instance, some approaches
propose modifying a DMP during execution to fit a desired force profile, which is stored in an external memory buffer \cite{AbuDakka2015Adaptationmanipulationskills}. A PI-Admittance controller uses this stored force profile and the tracking error to adapt the trajectory \cite{AbuDakka2015Adaptationmanipulationskills, Wang2021RoboticImitation}. 
In contrast, \famp{} directly encodes the force profile within the MP parameter space, eliminating the need for external memory, sensors, or special control schemes.

Other methuds use a hybrid control scheme to modulate a DMP trajectory \cite{Steinmetz2015Simultaneouskinestheticteaching, Montebelli2015Handing}. A cartesian impedance controller executes the DMP's position trajectory while a second force controller enacts the desired forces. The hybrid controller switches between impedance and force control based on the measured external forces. This method enacts force along a single axis (z-direction) \cite{Steinmetz2015Simultaneouskinestheticteaching, Montebelli2015Handing} and cannot handle multiple demonstrations or a sequence of DMPs \cite{Manschitz2016Probabilisticdecompositionsequential}. \famp{}, however, is agnostic to the used controller, especially with regard to cartesian or joint space. \famp{} does not make any assumption regarding force direction or the number of force axes. Its probabilistic nature allows \famp{} to encode a set of demonstrations in a single MP.

Compliant Movement Primitives (CMP) use a DMP to encode the position trajectory and a torque primitive (TP) for the force profile \cite{Denisa2016LearningCompliantMovement}. The TP is learned by executing the same DMP trajectory with different controller gains, resulting in a distribution of associated torque trajectories. In contrast, \famp{} learns a single, joint representation of both position and force profile, eliminating the need for multiple trajectory executions, thus scaling better in a learning-from-demonstration context. CMPs aim for high tracking accuracy with \textit{low} contact forces \cite{Denisa2016LearningCompliantMovement, Petric2018SensorimotorCMP, Dou2022RobotSkillLearning}, whereas \famp{} models the required forces to achieve a task and will increase forces if necessary.

Other approaches propose using the demonstrated force profile to segment a long-horizon demonstration into multiple basic skills, which can then be sequenced to solve complex tasks \cite{Manschitz2016Probabilisticdecompositionsequential, Lioutikov2015ProbabilisticSegmentation, Manschitz2020SquentialForce, Lioutikov2020LearningAttributeGrammars}. in contrast, \famp{} does not segment trajectories, but uses a replanning mechanism to adapt the trajectory to the task. 
Interaction Primitives are an approach, which use replanning to adapt DMPs based on observations from a secondary agent's joint configuration \cite{amor2014interaction, Lai2022UserIntent, Abbatematteo2024ComposableInteraction}, such as a human collaborator tracked via an OptiTrack system. \famp{}'s event-based replanning mechanism is conditioned on the robot's measured forces, independent of external measurements, and relies solely on the robot's internal force sensors. During trajectory execution, the replanning mechanism is triggered once the measured forces exceed a task-dependent threshold.

\textbf{Real-World Manipulation Benchmarks.}
Robot learning in the real world faces challenges not always considered in simulated environments, such as automatic resets and noisy or delayed sensor inputs. Contact-rich manipulation tasks with complex surface collisions and task-relevant contact forces remain difficult to simulate, despite recent advances in robot simulators such as IsaacLab \cite{mittal2023orbit}. 
The NIST Assembly Task Boards \cite{Kimble2020BenchmarkingProtocolsEvaluating} offer a variety of complex and realistic assembly tasks, but require specialized industry hardware, presenting a higher threshold for parts availability and setup. 
FurnitureBench \cite{Heo2023FurnitureBenchReproducibleReal} consists of easily producible 3D printed parts and offers long-horizon furniture assembly tasks involving multiple skills. Each furniture piece requires the robot to pick up and attach parts. This benchmark is designed with a standardized robot and camera setup to facilitate accurate simulation. 
RGB-Stacking \cite{Lee2021pickplaceTackling} requires robotic agents to reason about the dynamics of stackable objects with diverse shapes. While the measured forces inform the agent about object dynamics, such as the center of gravity, the force profile is not a direct requirement for solving the task.

The POEMPEL suite is built from easily producible 3D printed parts,requiring only a few nuts and bolts for assembly. Its modular nature allows for different configurations, each uniquely identifiable via a baked-in encoding system. The peg-in-hole setup at the core of the task requires a specific amount of force to overcome initial resistance, with a tactile click upon successful plug insertion providing additional force feedback. The POEMPEL suite offers non-trivial failure modes for both insufficient and exceeding forces. While most benchmarks fail destructivelt when forces exceed limits, such as over-tightening screws in FurnitureBench, the POEMPEL socket allows the plug to be over-inserted, avoiding breakage.
\section{Force-Aware ProDMP}
\label{sec:method}
The objective of learning a force-aware movement primitive is to obtain a compact representation of both the position trajectory and the measured forces. The force-aware movement primitive specifies the expected forces for a generated trajectory and can generate a trajectory to apply a desired force. 
For instance, in a power plug insertion task, the robot end-effector must reach the correct position and orientation and apply enough force in order to achieve a secure connection. 
If the forces measured during trajectory execution deviate from the expected forces represented by the force-aware MP, the trajectory should adapt smoothly to correct this discrepancy.

ProDMP already provide some of the desired properties for force-aware MP. ProDMP capture the position statistics from a set of demonstrations and allow for smooth replanning of the trajectory at regular time intervals during execution \cite{Li2023ProDMPUnifiedPerspective}. Force-Aware ProDMP (\famp{}) build on ProDMP, but also encode the measured forces from the demonstrations. \famp{} can generate new trajectories conditioned on desired forces. During execution, \famp{} continuously compare the measured forces to the desired forces and can smoothly replan the trajectory as needed.

\subsection{Probabilistic Dynamic Movement Primitives}
Like DMP, ProDMP encodes a single position trajectory $\cvec{\lambda} = [y_t]_{t=0:T}$ with $T$ timesteps using a parameter vector $\cvec{\omega} = [\boldsymbol{w}, g]^\intercal$ with basis functions weights $\cvec{w}$ and goal attractor $g$. 

Similar to ProMP, ProDMP assumes the weight vectors of a set of demonstration trajectories follow a multivariate normal distribution $\cvec{\omega} \sim \mathcal{N}(\cmat{\omega} | \cvec{\mu}_{\omega}, \cmat{\Sigma}_{\omega})$. ProDMP can compute the trajectory distribution $p(\cmat{\Lambda})$ with full covariance over all time steps $0:T$ as

\begin{align}
\label{eq:famp}
    \begin{aligned}
    p(\cmat{\Lambda}; & \cmat{\mu}_{\omega}, \cmat{\Sigma}_\omega, \cvec{y_b}, \dot{\cvec{y}}_b) = \mathcal{N}(\cmat{\Lambda}|\cvec{\mu}_{\Lambda}, \cmat{\Sigma}_\Lambda), \\
    \cvec{\mu}_\Lambda &= \cvec{\xi}_1 \cvec{y}_b + \cvec{\xi}_2 \dot{\cvec{y}}_b + \cmat{H}^\intercal_{0:T} \cvec{\mu}_\omega, \\
    \cmat{\Sigma}_\Lambda &= \cmat{H}^\intercal_{0:T} \cmat{\Sigma}_\omega \cmat{H}_{0:T} + \sigma^2_n \cmat{I}, 
\end{aligned}    
\end{align}

with initial robot states $\cvec{y}_b, \dot{\cvec{y}}_b$, their coefficients $\xi_k(\cvec{y}, \dot{\cvec{y}}, \cvec{y}_b, \dot{\cvec{y}}_b)$ and a combination $\cmat{H}_{0:T}$ of block-diagonal basis function matrices. The covariance matrix $\cmat{\Sigma}_\Lambda$ captures the temporal and inter-DoF correlations of the underlying trajectories through the basis function combination $\cmat{H}_{0:T}$.
For full details, please refer to the original 2023 RA-L paper \cite{Li2023ProDMPUnifiedPerspective}.

Thus ProDMP combines the strengths of DMP and ProMP, offering smooth trajectories that can adapt to different initial conditions and support the
combination and blending of multiple trajectories \cite{Li2023ProDMPUnifiedPerspective}.

\subsection{Force Extension}
\famp{} encodes a robot trajectory with $D$ position DoFs and a $F$-dimensional force vector in a ProDMP.
For example, a 7 DoF robot arm like the Franka Emika has $D=7$ joint positions and $F=7$ joint torque measurements. \famp{} makes no assumption on the relationships between the dimensionalty of position DoFs and force vectors, thus supporting Cartesian space ($D=F=3$) or even configurations where not every joint has an associated force sensor ($D \neq F$).

\famp{} fits a ProDMP to the $D$-dimensional position trajectory with weights $[\cvec{\omega}_d]_{d=0:D}$ and the $F$-dimensional force trajectory with weights $[\cvec{\omega}_f]_{f=D+1:D+F}$. This changes the dimensionality of the learned trajectory distribution $p'(\cmat{\Lambda})$ and its parameters $\cvec{\mu}'_\Lambda, \cmat{\Sigma}'_\Lambda$ to $D+F$ dimensions, instead of the $D$ dimensions of a strictly positional ProDMP. \famp{}'s extended covariance matrix $\cmat{\Sigma}'_\Lambda$ thus captures the temporal, inter-DoF and position-force correlations.
During trajectory execution, the robot follows the position trajectory generated by \famp{}, i.e., the first $D$ dimensions of the generated trajectory.  

Like ProMP and ProDMP, \famp{} allow for partial conditioning on specified DoFs. At conditioning timestep $t$, all values of the block-diagonal combination matrix $\cmat{H}_t$ are set to $0$, except for the diagonal values $(\cmat{H}_t)_{(c,c)}$ for all specified conditioning DoFs $c$. The partially conditioned matrix $\cmat{H}'_t$ is then used with \autoref{eq:famp} to generate a conditioned trajectory.

\begin{figure}
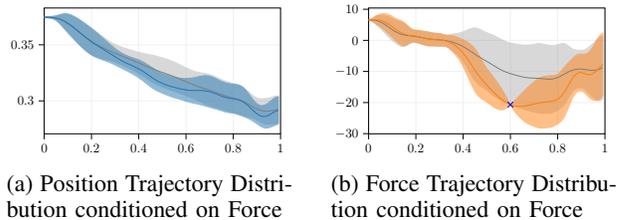

        \centering
        \begin{minipage}[t]{0.44\linewidth}
            \centering
            \resizebox{\textwidth}{!}{\input{plots/pos_conditioned}}
            \subcaption{\small{Position Trajectory Distribution conditioned on Force}}\label{fig:famp-pos-traj-cond}
        \end{minipage}
        \hspace{0.015\textwidth}
        \begin{minipage}[t]{0.44\linewidth}
            \centering
            \resizebox{\textwidth}{!}{\input{plots/force_conditioned}}
            \subcaption{\small{Force Trajectory Distribution conditioned on Force}}\label{fig:famp-force-traj-cond}
        \end{minipage}
        \caption{Position (blue) and Force Trajectory Distributions (orange) with mean and standard deviation. as generated by \famp{}. The gray distribution represents the corresponding unconditioned trajectory distribution. The colored distributions are conditioned at $t=0.6$ with desired force at $-20$. The position trajectory (blue) is implicitly altered through the correlation between position and force.}
        \label{fig:famp-trajectories}
\end{figure}

ProMP and ProDMP can only be conditioned on a desired position. \famp{}, however, can be conditioned on a desired position, desired force or both by selecting the conditioning dimensions $c$ from either the $D$ position or $F$ force dimensions. \autoref{fig:famp-trajectories} illustrates how the generated position and force trajectory distribution change when partially conditioned on only the desired force, due to the correlations captured by the covariance matrix $\cmat{\Sigma}'_\Lambda$.

\subsection{Force-Based Replanning}
\famp{} can capture force profiles from trajectories and be conditioned to enact desired forces. Like DMP and ProDMP, \famp{} can adapt the trajectory during execution. For example, when an external computer vision system detects changes in the environments, it can generate a new trajectory starting from the robot's current state.

\famp{} can also replan without relying on any external system, simply by measuring the forces acting on the robot. \famp{} computes at each timestep $t$

\begin{equation}
    \delta < \sum_f^F |(\cmat{\Lambda}_f)_t - (\tau_f)_t|,
\end{equation}

i.e., if the sum of absolute differences between the expected forces $\cmat{\Lambda}_t$, as encoded in the \famp{} trajectory, and the measured forces $\tau_t$ over all $F$ force dimensions exceeds a certain threshold $\delta$.
If this occurs, \famp{} generates a new trajectory conditioned on the measured forces using the partial conditioning described earlier. \famp{} dynamically selects the dimensions for conditioning. First, all force dimensions are ranked by their deviation from the expected value. The top $n$ dimensions are then selected for conditioning, covering at least $50 \%$ of the total difference. This ratio and the threshold $\delta$ are chosen empirically.

Because \famp{} continuously monitors all force dimensions, it can directly adapt to abrupt changes in the environment. This event-based replanning contrasts sharply with existing methods like ProDMP, which require either external replanning, or interval-based replanning at specific timesteps. \famp{}'s replanning is also agnostic to joint-based or cartesian control, as it makes no assumption regarding the nature of the force dimensions.

Even though \famp{} supports ``true'' replanning as designed for DMP and ProDMP, our experiments showed the best results when blending between the old and newly conditioned trajectory at a future timestep $t_{\mathrm{mix}}$.
The blending operation, which is also supported by ProMP and ProDMP, uses the sigmoid function $\sigma(x) = (1 + e^{-x})^{-1}$.
The resulting, desired trajectory $\Lambda_{\mathrm{des}}$ is calculated as a mixture between the old and newly conditioned trajectories according to
\begin{equation}
    \Lambda_{\mathrm{des}} = \Lambda_{\mathrm{old}} \cdot \sigma(-\gamma (t-t_{\mathrm{mix}})) + \Lambda_{\mathrm{cond}} \cdot \sigma(\gamma (t-t_{\mathrm{mix}})),
\end{equation}
where $\gamma$ is a factor defining the steepness of the sigmoid function (i.e. blending speed).

\begin{figure*}
        \centering
        \begin{minipage}[t]{0.22\linewidth}
            \centering
            \resizebox{\textwidth}{!}{\includegraphics{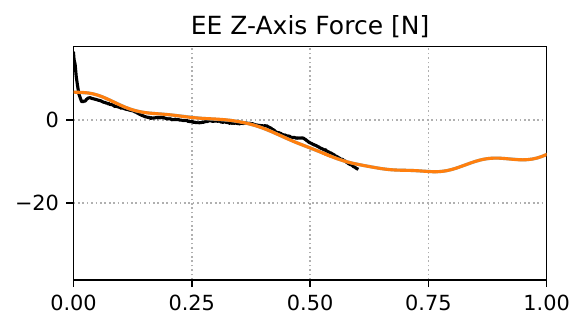}}
            \subcaption{\small{Trajectory Start}}\label{fig:famp-force-traj-replan-start}
        \end{minipage}
        \hspace{0.015\textwidth}
        \begin{minipage}[t]{0.22\linewidth}
            \centering
            \resizebox{\textwidth}{!}{\includegraphics{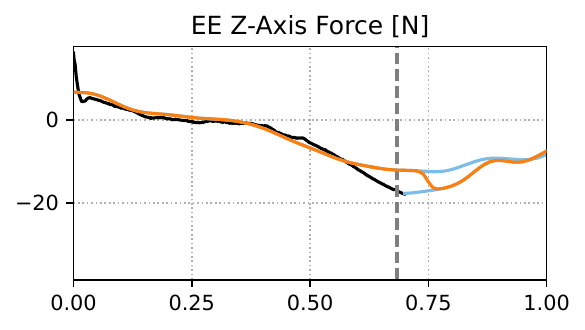}}
            \subcaption{\small{First Replanning Event}}\label{fig:famp-force-traj-replan-one}
        \end{minipage}
        \hspace{0.015\textwidth}
        \begin{minipage}[t]{0.22\linewidth}
            \centering
            \resizebox{\textwidth}{!}{\includegraphics{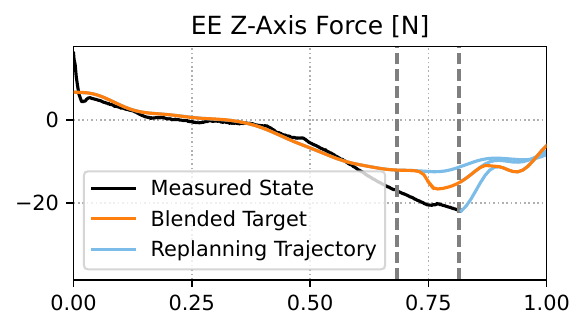}}
            \subcaption{\small{Second Replanning Event}}\label{fig:famp-force-traj-replan-two}
        \end{minipage}
        \hspace{0.015\textwidth}
        \begin{minipage}[t]{0.22\linewidth}
            \centering
            \resizebox{\textwidth}{!}{\includegraphics{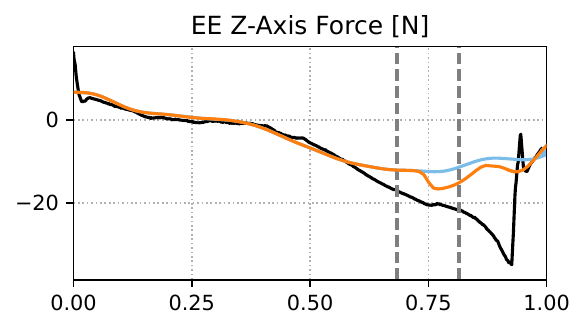}}
            \subcaption{\small{Completed Trajectory}}\label{fig:famp-force-traj-replan-end}
        \end{minipage}
        \caption{Z-axis force trajectory of \famp{} during vertical POEMPEL insertion experiment. (a) The robot initially follows the expected trajectory. (b) Upon encountering initial insertion resistance, the force exceeds the expected threshold, triggering replanning. (c) Continued resistance outside the expected range prompts a second replanning event. (d) The applied force eventually becomes sufficient to insert the POEMPEL plug into the socket, resulting in the tactile ``click''.}
        \label{fig:famp-trajectories-replanning}
\end{figure*}

\autoref{fig:famp-trajectories-replanning} illustrates the execution of a \famp{} trajectory on the POEMPEL setup with two replanning events. Initially, the robot state (black) follows the desired trajectory (orange). At two points, the measured forces exceed the expected range, prompting the generation of a newly conditioned trajectory (light blue). The desired trajectory is updated to blend the former target and the newly conditioned trajectory. Ultimately, the robot overcomes the resistance of the POEMPEL plug and socket. The resulting tacticle click is shown as a spike in the measured forces.

\section{POEMPEL Task Suite}
\begin{figure}
        \centering
        \begin{minipage}[t]{0.2\linewidth}
            \centering
            \includegraphics[width=\textwidth]{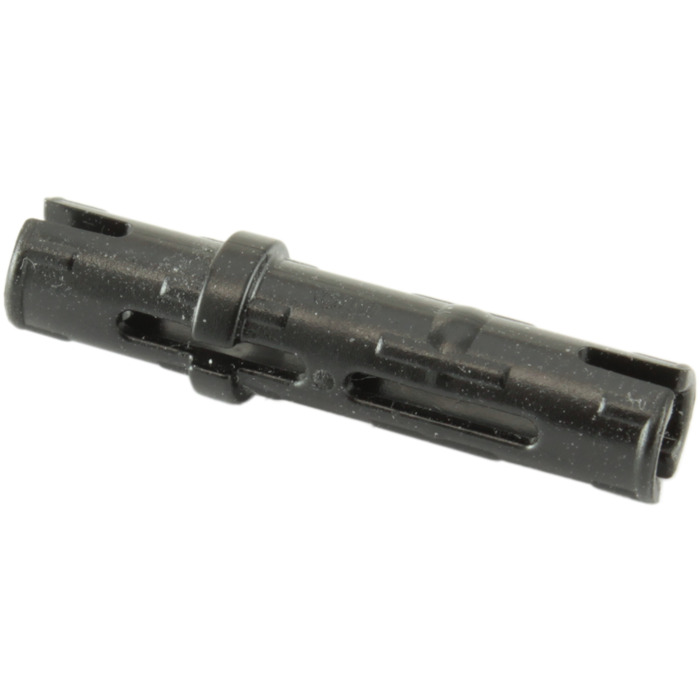}
            \subcaption{\small{Lego Item 6558}}\label{fig:lego_plug}
        \end{minipage}
        \hspace{0.015\textwidth}
        \begin{minipage}[t]{0.2\linewidth}
            \centering
            \includegraphics[width=\textwidth]{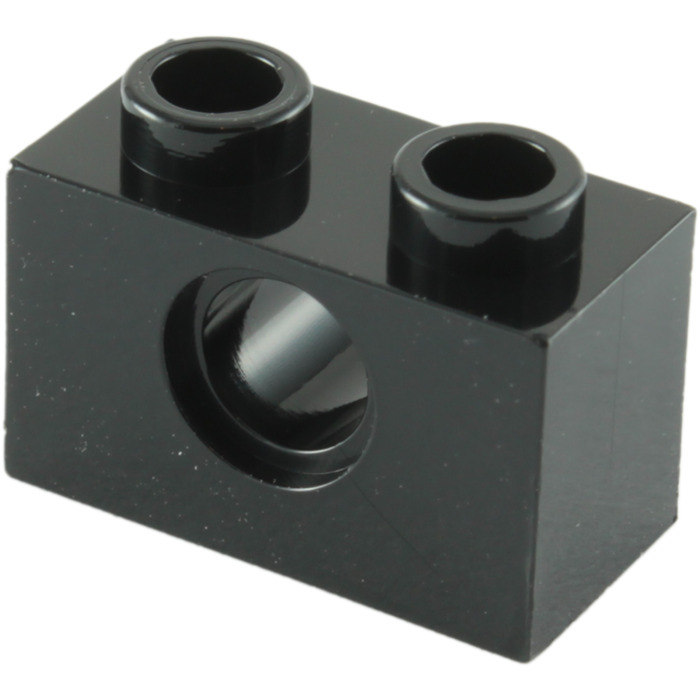}
            \subcaption{\small{Lego Item 3700}}\label{fig:lego_socket}
        \end{minipage}
        \hspace{0.015\textwidth}
        \begin{minipage}[t]{0.2\linewidth}
            \centering
            \includegraphics[width=\textwidth]{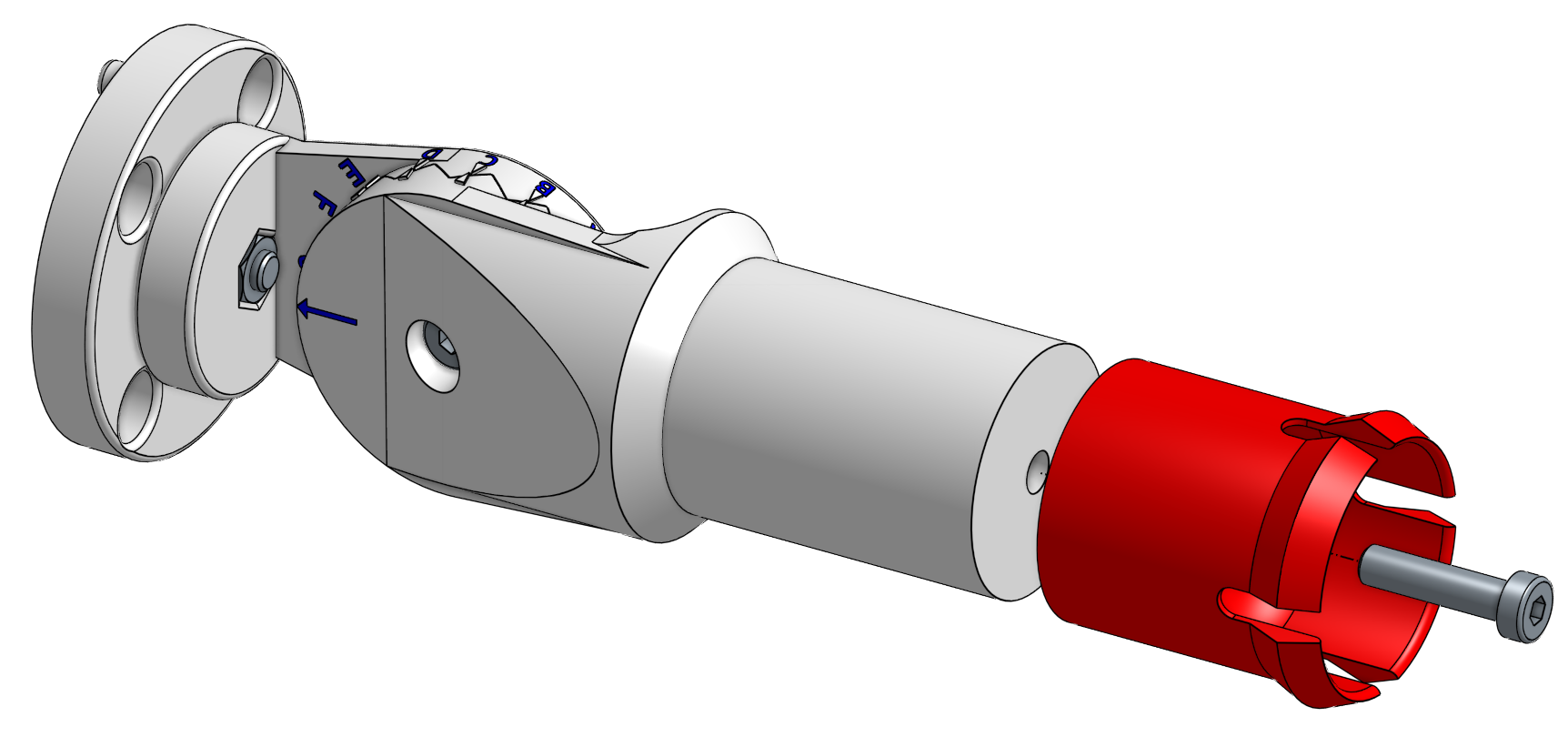}
            \subcaption{\small{POEMPEL plug}}\label{fig:poempel_plug}
        \end{minipage}
        \begin{minipage}[t]{0.2\linewidth}
            \centering
            \includegraphics[width=\textwidth]{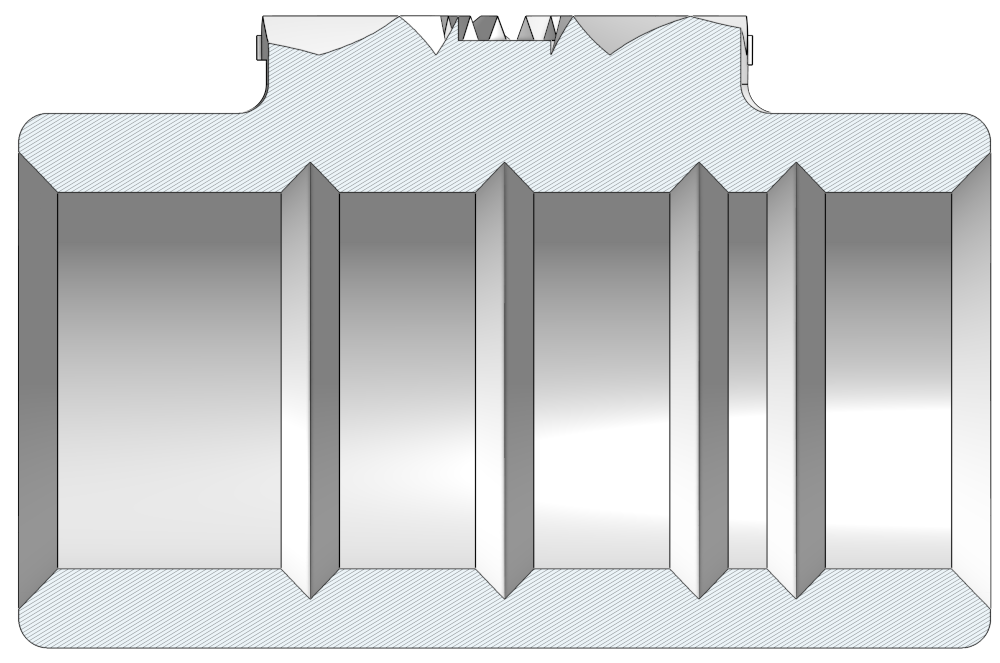}
            \subcaption{\small{POEMPEL socket section view}}\label{fig:poempel_socket}
        \end{minipage}
        \caption[]{Lego Parts inspiring the POEMPEL task suite and corresponding POEMPEL components.}
        \label{fig:legos}
\end{figure}

The ``Plastic Over-Engineered Movement Primitive Endeffector Link`` or POEMPEL (a german dialect word for ``plug'') is a novel task suite designed for force-aware robot manipulation.
Inspired by assembly toys like \textit{Lego Technic}, POEMPEL mimics industrial assembly tasks such as peg-in-the-hole or cable connectors, where the robot must successfully insert a cylindrical plug into a socket to ensure a secure connection.

\autoref{fig:legos} illustrates the main components of the POEMPEL setup and their \textit{Lego}-inspired design. The plug (\autoref{fig:poempel_plug}, in red) is fabriacted from flexible TPU 3D printing material. By varying the construction parameters of the POEMPEL plug, such as wall thickness or printing material, different degrees of stiffness can be achieved. Since the plug must be slightly compressed during insertion, the varying stiffness levels also alter the required insertion force. 

The POEMPEL socket is designed with slots at different depths (\autoref{fig:poempel_socket}), allowing for tasks with varying insertion positions. A successful plug insertion produces a tactile click, which can be measured as a spike in the force profile (see \autoref{fig:famp-force-traj-replan-end}). This feature enables tasks to be defined not solely by absolute position but also by more ambiguous terms, such as ``insert the plug until the third click''.

\autoref{fig:poempel} shows the entire POEMPEL system mounted on a Franka Emika robot arm. In addition to the plug and socket, the POEMPEL system includes a plug holder (\autoref{fig:poempel_plug}, white) and socket holder, each with two additional, indexed DoF. This allows each part to be mounted in different orientations. The plug holder is directly mounted to the robot arm, excluding the gripper as a variable in the robot's task performance. The socket holder is mounted on a 3D-printed grid plate, ensuring that each position on the plate can be clearly identified and replicated. The modular construction allows the grid plates to adapt to size of the available work space. An additional mounting bracket and exchangeable feet ensure that the grid plates are always in the same relative position to the robot base.
Modifying position and orientation of the POEMPEL parts allows for different tasks, requiring the robot to reach various positions and apply forces in different directions.

Nearly all parts of the POEMPEL system are 3D-printed, with the exception of a few nuts and bolts required for assembly. The parts are constructed so that all DoF are uniquely identifiable, making it possible to describe an entire experiment configuration with a six-character hex string. The modular nature of the system allows for future extension, such as additional manipulation tools or obstacles mounted to the grid plates.

We provide open access to the 3D printing and assembly instructions\footnote{Please find POEMPEL instructions at \href{https://intuitive-robots.github.io/poempel/}{https://intuitive-robots.github.io/poempel/}} and encourage other labs to also use this task suite to test their force-aware robot manipulation methods.
\section{Experiments}
\label{sec:experiment}
These experiments demonstrate the advantages of \famp{} by comparing its performance with other MP formulations, including DMP, ProMP, ProDMP, and Cartesian Impedance controller (CIC) replay  of a demonstration. The experiments use a Franka Emika 7-DoF arm controlled with Polymetis \cite{Polymetis2021}, a robot control framework written for Pytorch. The evaluated tasks include several POEMPEL setups and the insertion of a common electrical power plug into a power socket.

\subsection{Data Collection}
To collect human demonstrations via kinesthetic teaching, a teleoperation framework with force-feedback is employed. The human operator directly moves the primary robot, while a replicant robot mirrors the movements using an end-effector-based impedance controller. Forces measured on the replicant robot joints are reflected on the primary robot, providing haptic feedback and control of the applied forces to the human operator. Since the replicant is not in gravity compensation mode and does not experience external forces induced by the human operator, replaying the recorded trajectory is unnecessary for collecting a valid force profile for MP encoding.

\subsection{POEMPEL Setups}
These are the variables for the various experiments:
\begin{itemize}
    \item \textit{POEMPEL Orientation:} The POEMPEL socket and plug orientation is either vertical or horizontal.
    \item \textit{Socket Position:} The vertical experiments use a fixed socket position. For the horizontal experiments, the socket is mounted at two different positions, altering the travel distance of the trajectory.
    \item \textit{Plug Stiffness:} Two different plugs with soft and firm stiffness levels are used, changing the amount of force required to successfully seat the plug in the socket.
\end{itemize}

\begin{figure}
    \centering
    \includegraphics[width=0.6\linewidth]{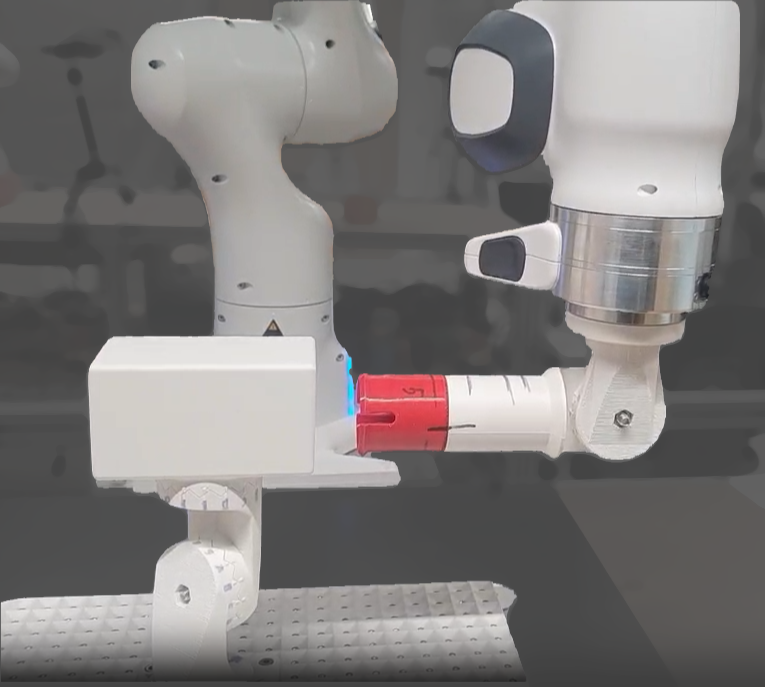}
    \caption{Horizontal Orientation of POEMPEL plug and socket on a Franka Emika 7-DoF robot arm.}
    \label{fig:horizontal}
\end{figure}

\autoref{fig:poempel} shows the vertical setup, and \autoref{fig:horizontal} depicts the horizontal setup. For each setup, seven human demonstrations are collected. Each experiment run consists of seven trajectory executions, either sampled from the MP trained on the entire demonstration distribution (ProMP, ProDMP, \famp{}) or by sampling and encoding a random demonstration (DMP, CIC).

\begin{table*}[]
\centering
\begin{tabular}{|l||l|l|l|l|l|}
 Experiment            & CIC     & DMP     & ProMP   & ProDMP  & \textbf{FA-ProDMP (Ours)} \\ \hline
Demonstration Replay  & \cellcolor{green!25}0.41 cm & \cellcolor{green!25}0.62 cm & \cellcolor{green!25}0.40 cm & \cellcolor{green!25}0.38 cm & \cellcolor{green!25}\textbf{0.13 cm}          \\
Vertical Adaptation   & 2.99 cm   & 3.05 cm   & 2.99 cm & 2.99 cm    & \cellcolor{green!25}\textbf{0.12 cm}          \\
Horizontal Adaptation & 3.72 cm   & 3.98 cm  & 3.54 cm    & 3.51 cm    & \cellcolor{green!25}\textbf{0.27 cm}  \\
Electr. Power Plug & 1.15 cm & 1.15 cm & 1.15 cm & 1.15 cm & \cellcolor{green!25}\textbf{0.45 cm}
\end{tabular}
\caption{Mean Position Error after trajectory execution. Successful trajectories are highlighted in green. Cartesian Impedance Control (CIC) replay, DMP, ProMP and ProDMP succeed only when the demonstrations match the testing setup. Only \famp{} can adapt to new situations through to event-based replanning, resulting in successful insertions.}
\label{tab:results}
\end{table*}

\subsection{POEMPEL Experiment Results}
Our experiments showed no difference in performance between Cartesian space and joint space control. For simplicity, only the results of the experiments with Cartesian control are presented in this paper.
\autoref{tab:results} provides an overview of the mean position error over seven runs.

\subsubsection{Demonstration Replay}
In the first experiment, all methods were trained on seven demonstrations collected with the soft POEMPEL plug in the vertical setting and tested on the same setup. As expected, all trajectories successfully inserted the POEMPEL plug into the socket when the testing setup parameters matched the training demonstrations, i.e., the plug stiffness remained unchanged. This was also true for training and testing with the firm POEMPEL plug or in the horizontal orientation.
As long as the demonstrations matched the testing setup, all methods solved the task.

\subsubsection{Vertical POEMPEL Force Adaptation}

In this experiment, the POEMPEL setup was orientated vertically. 14 demonstrations were provided: Seven with the soft POEMPEL plug and seven with the firm plug. All tests were executed with the firm POEMPEL plug, simulating an undetected change in the execution environment, such as variations in manufacturing tolerances.

Because the demonstration set was mixed and did not fully match the testing environment, DMP and CIC succeeded only when a matching demonstration was sampled; otherwise, they failed. The probabilistic methods ProMP and ProDMP always had access to the correct demonstrations as encoded in their learned trajectory distribution. However, since the methods learn the mean over the all the given demonstrations, they are susceptible to mode averaging and failed to insert the plug into the socket during testing time.

\famp{} adapted to the change in the testing environment. \autoref{fig:famp-trajectories-replanning} depicts the Z-Axis force trajectory of one example execution. \famp{} detected a mismatch between the expected (orange) and measured (black) force during runtime. This triggered two replanning events over the course of the trajectory execution. New trajectories were generated by conditioning on the measured forces, and the blending mechanism allowed \famp{} to smoothly transition to the new trajectory. The spike in the measured state marks the tactile click when the robot successfully seated the POEMPEL plug. This demonstrates how \famp{} adapts the trajectory by progressively increasing the desired forces until the task is completed.

\subsubsection{Horizontal POEMPEL Adaptation}
The final POEMPEL experiment involved the horizontal POEMPEL orientation. The demonstration dataset consisted of 14 trajectories with the socket at various positions along the workspace's Y-Axis. These trajecotries varied not only in their approach to the socket but also in total travel distance and the final Y position of the robot end-effector upon plug insertion. This task simulates the variance in the grasping of manipulation objects, such as when the robot grabs a cable connector at slightly different relative positions to the cable end, thereby altering the total movement distance. 

Detecting such a change in the robotic grasp remains challenging. Even though a calibrated vision system would be able to detect the socket's new position, without any external sensors, DMP and CIC could only solve the task when the randomly selected demonstration precisely matched the experimental setup. ProMP and ProDMP exhibited mode averaging, which prevented the successful insertion of the POEMPEL plug into the socket.

\begin{figure}
    \centering
    \includegraphics[width=0.8\linewidth]{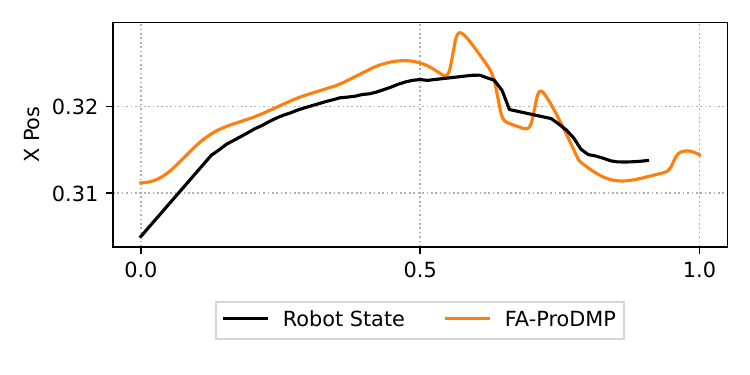}
    \caption{\famp{} X-Axis position trajectory (orange) and actual robot state (black) during horizontal POEMPEL adaptation experiment. The steps in the trajectory result from \famp{}'s replanning to new trajectories as the robot corrects for a slight misalignment in the angle between POEMPEL plug and socket.}
    \label{fig:haf-famp}
\end{figure}

\famp{} succeeded in this task by relying solely on the feedback from the internal force sensors. \autoref{fig:haf-famp} illustrates the X-Axis position trajectory of one example. The \famp{} trajectory displays several replanning events, indicated by steps in the trajectory line. The robot performed slight adjustments to account for minor misalignments between the POEMPEL plug and socket.

\subsection{Electrical Power Plug Task}
\begin{figure}
    \centering
    \includegraphics[width=0.7\linewidth]{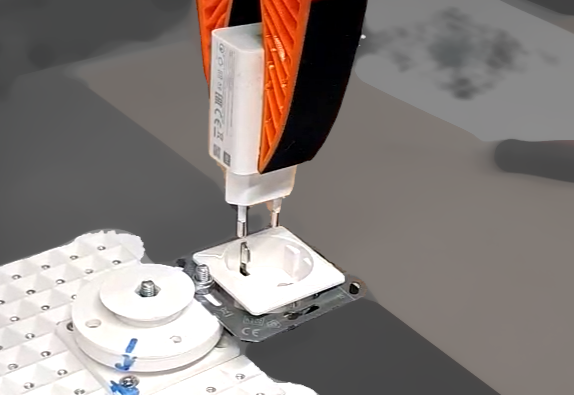}
    \caption{Experiment Setup with EU type F power socket and EU type C power plug. The POEMPEL grid base ensures a fixed setup and repeatability.}
    \label{fig:power-plug}
\end{figure}

The POEMPEL setup is designed to minimize uncertainty in task execution and reduce reliance on the installed gripper's grasping capabilities. To validate this setup as a proxy for real-world manipulation tasks, an experiment was conducted with an EU type C power plug and EU type F power sockets with varying plug-in resistances (see \autoref{fig:power-plug}). The difference in plug-in resistance is undetectable by an external vision system. All methods were trained on seven human demonstrations with the low-resistance power socket and a single demonstration with the high-resistance socket. 

Except for \famp{}, all other methods failed to insert the power plug into the high-resistance socket. Only \famp{} could increase the desired forces using event-based replanning and successfully insert the power plug into the power socket. This demonstrates how \famp{} can solve real-world tasks and adapt the force profile based on only a single additional demonstration.

\section{Conclusion \& Future Work}
\label{sec:conclusion}
\famp{}, a force-aware movement primitive based on ProDMP, effectively represents a distribution of desired positions and forces learned from demonstrations. It generates new trajectories conditioned on desired positions, forces, or both, and is compatible with both joint and Cartesian control.
Through replanning and blending, \famp{} adapts trajectories during execution based on measured forces, as demonstrated in multiple real-robot experiments using the newly proposed POEMPEL task suite and an electrical power plug experiment. 

\famp{} currently faces limitations regarding safety with human collaborators. As a non-compliant system, it may increase output force to counter unexpected resistance. Future work aims to address this by integrating \famp{} with multi-modal sensors and deep learning policies for imitation learning. This integration would enable \famp{} to adapt robot force profiles in the presence of human collaborators, enhancing safety in human-robot interactions.

\bibliographystyle{IEEETran}
\bibliography{references}
\clearpage

\end{document}